\documentclass[journal]{IEEEtran}
\usepackage{amsmath,amsfonts,amssymb}
\usepackage{algorithmic}
\usepackage{algorithm}
\usepackage{array}
\usepackage[caption=false,font=normalsize,labelfont=sf,textfont=sf]{subfig}
\usepackage{textcomp}
\usepackage{stfloats}
\usepackage{url}
\usepackage{verbatim}
\usepackage{graphicx}
\usepackage{comment}
\usepackage{cite}
\usepackage{xspace}
\usepackage{xcolor}
\usepackage{color}
\usepackage{bbding}
\usepackage{lineno}
\usepackage{booktabs}
\usepackage{multirow}
\usepackage{arydshln}
\usepackage{enumitem}
\usepackage{hyperref}

\definecolor{MyDeepGreen}{RGB}{0,150,0} 

\definecolor{MyDeepBlue}{RGB}{50,150,240} 

\begin{document}

\title{Low-light Image Enhancement with Retinex Decomposition in Latent Space}

\author{Bolun Zheng, Qingshan Lei, Quan Chen, Qianyu Zhang, Kainan Yu, Xu Jia and Lingyu Zhu
\thanks{Bolun Zheng, Qingshan Lei, Qianyu Zhang and Kainan Yu are with the School of Automation, Hangzhou Dianzi University, Hangzhou 310018, China~(e-mail:blzheng@hdu.edu.cn; 241060071@hdu.edu.cn; zqqqyu@163.com; 244060059@hdu.edu.cn).}
\thanks{Quan Chen is with the College of Artificial Intelligence, Jiaxing University, Jiaxing 314001, China~(e-mail:chenquan@alu.hdu.edu.cn).}
\thanks{Xu Jia is with the School of Artificial Intelligence at Dalian University of Technology, Dalian 116000, China~(e-mail:jiayushenyang@gmail.com).}
\thanks{Lingyu Zhu is with the Department of Computer Science at City University of Hong Kong, Hong Kong, China~(e-mail:lingyzhu-c@my.cityu.edu.hk).}
\thanks{Quan Chen is the Corresponding Author.}
}

\markboth{Journal of \LaTeX\ Class Files,~Vol.~14, No.~8, August~2021}%
{Shell \MakeLowercase{\textit{et al.}}: A Sample Article Using IEEEtran.cls for IEEE Journals}


\maketitle

\begin{abstract}
Retinex theory provides a principled foundation for low-light image enhancement, inspiring numerous learning-based methods that integrate its principles.
However, existing methods exhibits limitations in accurately decomposing reflectance and illumination components.
To address this, we propose a Retinex-Guided Transformer~(RGT) model, which is a two-stage model consisting of decomposition and enhancement phases.
First, we propose a latent space decomposition strategy to separate reflectance and illumination components. 
By incorporating the log transformation and 1-pixel offset, we convert the intrinsically multiplicative relationship into an additive formulation, enhancing decomposition stability and precision.
Subsequently, we construct a U-shaped component refiner incorporating the proposed guidance fusion transformer block.
The component refiner refines reflectance component to preserve texture details and optimize illumination distribution, effectively transforming low-light inputs to normal-light counterparts.
Experimental evaluations across four benchmark datasets validate that our method achieves competitive performance in low-light enhancement and a more stable training process.
\end{abstract}


\section{Introduction}
\IEEEPARstart{L}{ow-light} image enhancement serves as a critical yet challenging task in computer vision, one that underpins a wide array of visual tasks including autonomous driving~\cite{wang2022sfnet} and object detection~\cite{chen2025wxsod}.
Images captured under low-light conditions often suffer from prominent degradations such as severe color distortion, amplified noise and compromised detail visibility. 
These impairments severely hinder human visual perception but also considerably degrade the performance of various downstream visual tasks.
Conventional approaches, such as histogram equalization and gamma correction, generally exhibit limited robustness in handing real-world noise. 
Another line of methods, grounded in the Retinex theory, decomposes an image into reflectance ($R$) and illumination ($L$) components by assuming spatially smooth and slowly varying illumination. 
However, these Retinex-based methods focus on illumination estimation and tend to overlook the influence of noise. 
Abrupt illumination variations under extreme low-light conditions can further give rise to severe noise amplification or local color distortion.

\begin{figure}[!t]
\centering
\includegraphics[width=1.0\linewidth]{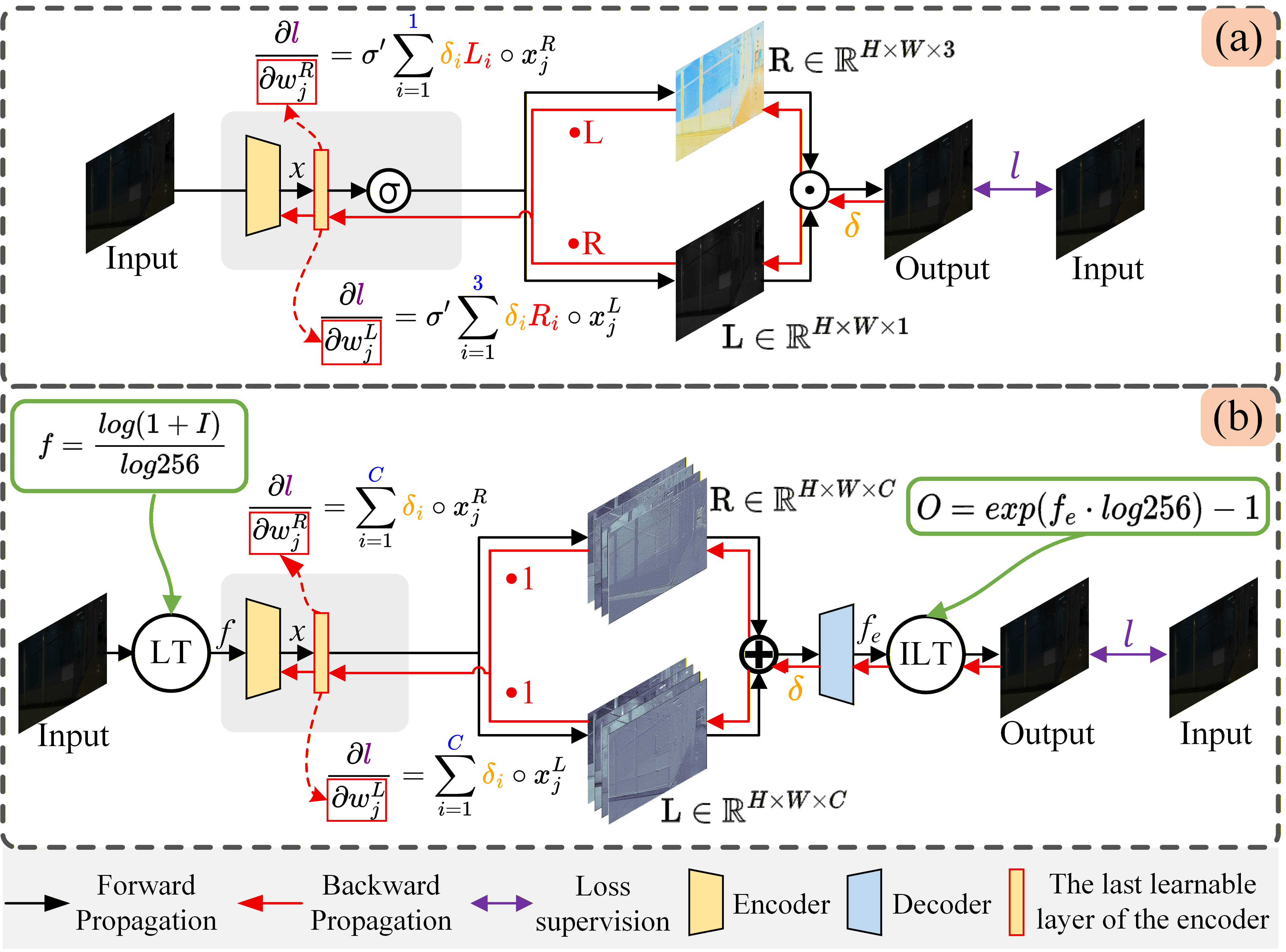}
\caption{Comparison between the typical Retinex decomposition strategy and our method in deep learning frameworks.}
\label{fig:first-img}
\end{figure}

With the vigorous advancement of deep learning, numerous low-light enhancement models based on convolutional neural networks~(CNNs) have emerged.
The mainstream design paradigm for such models is to learn the brightness mapping between low-light images and their normal-light counterparts. 
By treating details and noise as equally important information, these methods are prone to inducing color distortion and noise amplification. 
Another category of models incorporates Retinex theory into network design, typically consisting of an illumination estimation module followed by a refinement module.
For the image decomposition process, as shown in Fig.~\ref{fig:first-img}(a), these methods decompose an input image into 1-channel illumination component and 3-channel reflectance component in the color domain

Nevertheless, these methods exhibit two inherent limitations:
(1) The representational capability of the $R$ and $L$ components generated within the color domain is limited. During the camera imaging process, real-world illumination and sensor noise undergo nonlinear coupling, yet the linear model relying solely on the product of $R$ and $L$ struggles to model such complex scenarios.
(2) During the image reconstruction through element-wise multiplication, $R$ and $L$ components are integrated into the backpropagation process. Estimation errors of components in challenging regions directly undermine the training stability of the model~\cite{ecmamba,RetinexNet}.

To address these issues, we propose a Retinex-guided Transformer~(RGT) model consisting of a decomposition stage and an enhancement stage.
In the decomposition stage, we design a latent-space decomposition strategy, as shown in Fig.~\ref{fig:first-img}(b). 
Unlike conventional color-space decoupling methods, this strategy maps the input image into a latent feature space for nonlinear decomposition. 
A dual-branch decomposition network built upon Transformer blocks is employed, leveraging the global modeling ability of Transformers to improve the separation accuracy of reflectance and illumination components. 
Meanwhile, a logarithmic transformation is applied to the image in the color domain, converting the inherent multiplicative relationship between $R$ and $L$ into an additive one. 
This conversion effectively mitigates mutual interference during training and enhances optimization stability. 
It should be noted that a 1-pixel offset is introduced prior to the logarithmic transformation to prevent numerical instability caused by extremely low pixel values in dark regions.
In the enhancement stage, we design a Component Refiner~(CR) to separately enhance the $R$ and $L$ components.
We posit that average pooling operations are effective in preserving texture details while suppressing noise, whereas max pooling operations better retain illumination information.
Building on this insight, we develop a Guidance Fusion Transformer Block (GFTB) and stack it within a U-shaped architecture to construct the CR module.
Experimental results on four public datasets~(\textit{i.e.}, LOLv2~\cite{LOLv2}, HDR~\cite{hdr}, SDSD~\cite{sdsd}, SICE~\cite{SICE}) validate the effectiveness of our RGT model. 
For instance, on the HDR test set, RGT outperforms RetinexFormer by 0.79 dB in PSNR.
Extensive ablation studies demonstrate that the latent-space decomposition strategy achieves more effective decoupling of illumination and reflectance, while also promoting greater stability throughout the training process.

\begin{itemize}
\item We propose a retinex-guided transformer~(RGT) model that employs a two-stage paradigm of decomposition and enhancement to enhance low-light images.
\item We propose a latent space decomposition strategy to reconstruct images in the feature domain in an additive manner, ensuring component quality and model training stability.
\item We design a feature-guided transformer block that leverages image structural priors to refine image enhancement details.
\item Extensive experiments on four public datasets demonstrate the superior performance of the RGT model.
\end{itemize}

The rest of this paper is structured as follows. We first introduce related works in Section~\ref{Related Works}. In Section~\ref{Methods}, we present the proposed RGT in detail. The experimental results are presented in Section~\ref{Experiments and Results}. Finally, the concluding remarks are drawn in Section~\ref{Conclusions}.

\section{Related Work}\label{Related Works}
\subsection{Low-light Image Enhancement}
\textbf{Traditional Methods.} 
Numerous approaches have been developed in traditional low-light image enhancement. An elementary strategy is the direct and uniform amplification of image intensity. This global operation, however, frequently results in over-enhanced or under-enhanced regions. To alleviate these issues, techniques such as histogram equalization~\cite{clahe} and grayscale transformation methods~\cite{gamma2} have been widely adopted. Histogram equalization improves visibility by stretching the dynamic range of pixels in dark areas, thereby enhancing both contrast and brightness. Grayscale transformations, including gamma correction and logarithmic mapping, adjust each pixel independently without considering spatial contextual information. Although computationally efficient, these methods often amplify noise or exhibit limited capability in local adaptation. Another line of work~\cite{llie-dehaze} builds on dehazing theory, where the low-light image is inverted to mimic a hazy scene. Conventional dehazing algorithms are then applied, and the output is inverted back to produce the enhanced image. A fundamental limitation of these approaches is their strong dependence on the image dehazing process, which tends to introduce amplified noise and color artifacts.

\textbf{Deep Learning Methods.} 
The rapid development of deep learning has brought substantial performance gains to various image processing domains, leading to its widespread use in tasks such as image dehazing~\cite{dehaze-zhao2025transdehaze}, super-resolution~\cite{isr-umirzakova2024medical}, deblurring~\cite{deblur-xiang2025deep}, and low-light image enhancement~\cite{aaai1,iclr2,eccv2,eccv4,eccv3}. Further expanding the methodology spectrum, a number of recent approaches also incorporate diffusion models into the computational pipeline~\cite{iclr1,tip1,eccv5,iccv1}. In the context of low-light enhancement, most deep learning techniques rely on supervised training with paired datasets. A widely adopted paradigm employs convolutional neural networks~(CNNs) to learn an end-to-end mapping that directly transforms a low-light input into its normally-light counterpart~\cite{EEMEFN,jiang2024joint-cnn,luo2025low-cnn,zainab2025light-cnn,jiang2024mutual-cnn}.
One representative example is LLNet~\cite{lore2017llnet}, which implements a stacked sparse denoising autoencoder with a CNN-based backbone. This architecture is tailored to perform simultaneous brightness enhancement and noise suppression while effectively capturing multi-scale image characteristics. 
To address the challenge of balancing efficiency and performance, Li~\textit{et al.}~\cite{li2025wv} propose an image enhancement method based on 3D lookup tables. This method achieves competitive image quality by leveraging the high efficiency inherent to lookup tables.
In general, these methods optimize parameters via loss functions defined within a paired-supervised CNNs framework. 
These methods share a common limitation that their integration of physical image formation principles remains insufficient. 
This limitation reduces the interpret-ability of the models, but also frequently leads to suboptimal outcomes such as local over-enhancement or under-enhancement.

\textbf{Retinex-Based Methods.}
The Retinex theory has served as a foundational framework for low-light image enhancement since its inception~\cite{retinex}. Early implementations of this theory include the Single Scale Retinex (SSR) method~\cite{ssr}, which estimates a smooth illumination component through a single-scale Gaussian blur kernel. The reflection component is subsequently derived from the estimated illumination map $L$ and the original input, with the resulting $R$ component taken as the enhanced output. A key limitation of SSR lies in its use of a fixed filtering process that lacks learnable parameters, frequently leading to unnatural appearances and noticeable color distortion. To address these issues, later Multi-Scale Retinex (MSR) techniques~\cite{msr} were developed, which combine outputs from multiple SSR instances via weighted averaging. Although these methods improve the overall enhancement quality, they still exhibit persistent artifacts such as grayish tones and loss of fine details. With the progress of deep learning, Retinex theory has been increasingly incorporated into neural network designs~\cite{tmm1,sun2025di-retinex,xu2024cretinex,hu2025low,zhao2025retinex}, offering a physically motivated guidance mechanism. Most deep Retinex-based methods adopt a two-stage structure, beginning with the decomposition of the input into reflectance $R$ and illumination $L$, followed by a dedicated enhancement phase. One branch of methods, inspired by conventional SSR, focuses solely on estimating the illumination component $L$. As an example, Wang \textit{et al.}~\cite{deepupe} designed a CNN to predict $L$, after which the enhanced image is generated by dividing the original input by the estimated $L$. This paradigm, however, frequently fails to account for complex noise patterns, color inaccuracies, and physical constraints of the imaging process, often resulting in visible artifacts under demanding conditions. Another line of research jointly estimates both $R$ and $L$ in the decomposition stage. A prominent implementation, RetinexNet~\cite{RetinexNet}, utilizes a convolutional network to decompose the image, applies independent denoising and brightness enhancement to each component, and finally reconstructs the image through multiplicative fusion. A notable drawback of this design is that the multiplication-based recombination couples both $R$ and $L$ in the gradient computation during backpropagation, potentially destabilizing the training process. Furthermore, decomposition conducted in the original image space often results in incomplete separation between $R$ and $L$, which in turn constrains the final enhancement performance.  

\subsection{Transformer}
The Transformer architecture, originally developed for machine translation~\cite{transformer}, has seen growing adoption in computer vision due to its powerful global modeling capabilities. 
It has been successfully applied to diverse vision tasks including object detection~\cite{DETR,Swin}, image segmentation~\cite{segformer}, and low-light image enhancement~\cite{lytnet,deformer,cvpr1,eccv1,aaai2,wang2025rethinking,tu2025fourier,lin2025geometric}. 
In the domain of low-level vision, the U-shaped Transformer design has become particularly prevalent. 
Among these, Uformer~\cite{wang2022uformer} introduced a purely Transformer-based U-shaped encoder-decoder that employs skip connections to integrate multi-scale features, though its computational requirements remain substantial.
To address efficiency limitations, Restormer~\cite{zamir2022restormer} proposed the Multi-Dconv Head Transposed Attention (MDTA) and Gated-Dconv Feed-Forward Network (GDFN), which significantly reduce computational complexity while maintaining restoration performance. 
For the specific task of low-light enhancement, LLformer~\cite{llformer} introduced an axis-based multi-head self-attention mechanism that modifies the standard Transformer block to achieve lower computational cost while delivering competitive enhancement quality. Further extending this direction, several studies have combined Retinex theory with Transformer architectures. For instance, Liu~\textit{et al.}~\cite{liu2024efficient} first estimate an illumination component $L$ following Retinex principles to produce an initial brightened image. This intermediate output is then processed by a Multiscale Degradation Estimation Network built with customized Transformer blocks to further improve visual quality. Although the self-attention mechanism inherently captures long-range dependencies and supports global illumination adjustment, its undirected attention allocation often leads to inadequate reconstruction in severely degraded image regions. To mitigate this, Retinexformer~\cite{retinexformer} incorporates externally generated guidance signals through multiplication with the value vector, thereby enhancing restoration. However, this strategy still does not fully integrate guidance into the query–key interaction, which limits finer control over attention distribution.

\section{Methodology}\label{Methods}
\begin{figure*}[!t]
\centering
\includegraphics[width=1.0\linewidth]{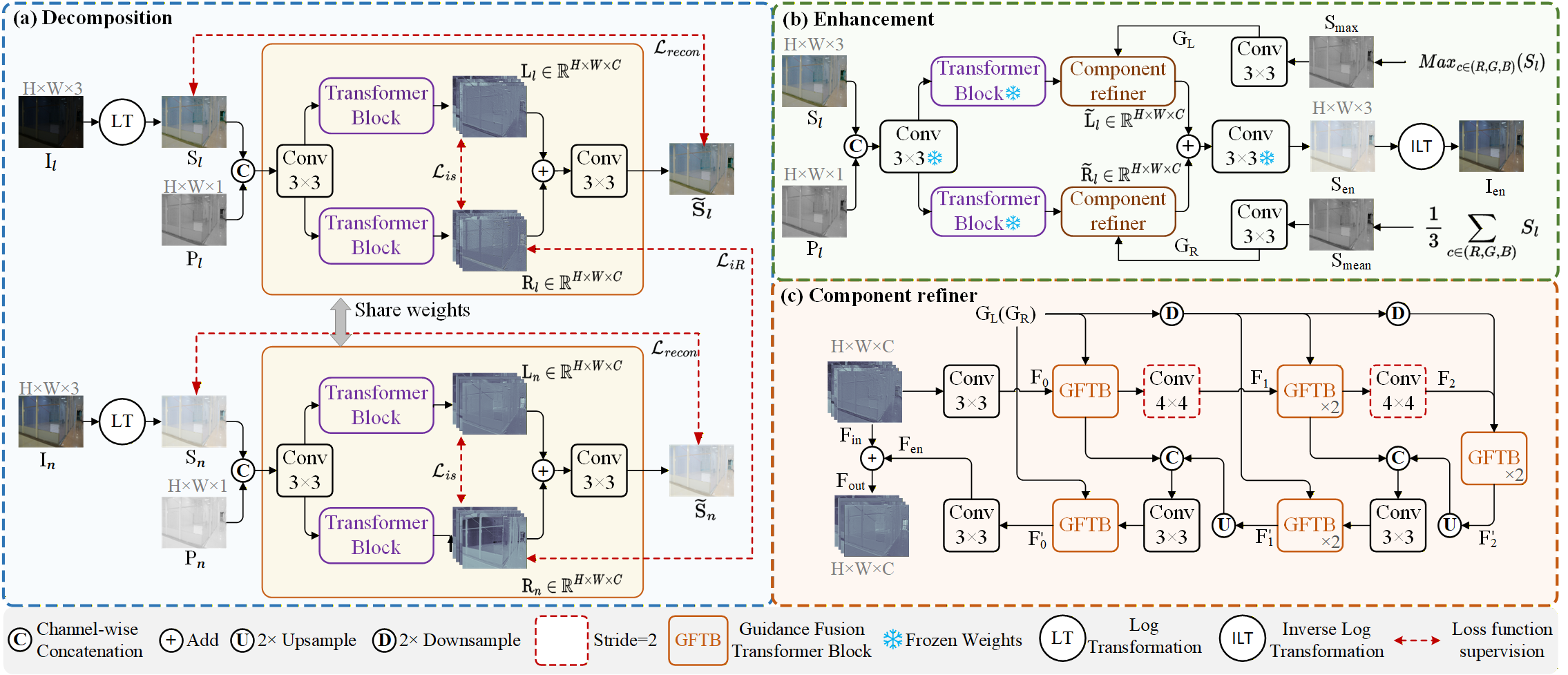}
\caption{The overview of the RGT model. 
(a) Decomposition takes a paired low-light/normal-light image as input during training to decompose the image into $R$ and $L$ components. 
(b) RGT extracts $R$ and $L$ using the pre-trained decomposition network (frozen parameters) and enhances $R$ and $L$ separately via the Enhancer network (non-shared parameters). 
Then, reconstructing the final enhanced result by combining processed $R$ and $L$.}
\label{framework}
\end{figure*}

As shown in Fig.~\ref{framework}, the proposed RGT model consists of two phases, namely image decomposition~(Fig.~\ref{framework}(a)) and enhancement (Fig.~\ref{framework}(b)).
We first review the Retinex theory, followed by the introduction the proposed latent space decomposition strategy.
Next, we present the component refiner used for low-light enhancement.
Finally, the loss function is described.

\subsection{Preliminary}
Based on Retinex theory~\cite{retinex}, a color image $I\in \mathbb{R}^{H\times W\times 3}$ can be decomposed into a reflectance map $R\in \mathbb{R}^{H\times W\times 3}$ and an illumination map $L\in \mathbb{R}^{H\times W\times 1}$:
\begin{equation}
I=R\odot L,
\label{eq1}
\end{equation}
where $\odot$ denotes an element-wise multiplication.

Current deep Retinex-based models~\cite{RetinexNet,ecmamba} universally employ dual-branch networks to estimate reflectance $R$ and illumination $L$ components in the color domain.
These components are supervised by reconstructing the input image via element-wise multiplication~(see Eq.~(\ref{eq1})).
Since this decomposition process is entirely data-driven, the gradient propagation of this process exerts a critical influence on component quality.
Considering the gradient for learnable parameters layers dedicated to estimating $R$ and $L$, their gradient back-propagation can be expressed as:
\begin{equation}
\frac{\partial l}{\partial w^R_j}=\delta^R_j\circ x^R_j,\quad \frac{\partial l}{\partial w^L_j}=\delta^L_j\circ x^L_j
\label{eq2}
\end{equation}
where $\delta^R_j$ and $\delta^L_j$ denote the local gradient of the $j$-th variable~(\textit{i.e.}, $w^R_j$ and $w^L_j$) in the convolution layer for estimating $R$ and $L$, $x^R_j$ and $x^L_j$ are the corresponding inputs, and $l$ denotes the back-propagation loss. 
Then the local gradient $\delta^{R/L}_j$ can be calculated using the chain rule as:
\begin{equation}
\delta^R_j=\sigma'\sum^m_{i=1}\delta_iL_i,\quad \delta^L_j=\sigma'\sum^m_{i=1}\delta_iR_i
\label{eq3}
\end{equation}
where $\sigma'$ denotes the derivative of the activation function, and $\delta_i$ denotes the local gradient of the $i$-th variable in the convolution layer for estimating $R$ and $L$. 
According to Eq.~(\ref{eq1}), $\frac{\partial I}{\partial R}=L$ and $\frac{\partial I}{\partial L}=R$. 
Therefore, the $R$ and $L$ are introduced into local gradient. 
For low-light image input, the values of $R$ and $L$ are generally small, leading to the slower convergence rate due to gradient scaling.
Besides, the decomposed $R$ and $L$ may exist abnormal values, leading to unbalanced parameters update.

Furthermore, previous methods perform image decoupling in color space and consider the combination process of $R$ and $L$ as a simple linear relationship. 
It interfere with the expression ability of the components. 
In addition, these methods often apply clip operations to $R$ and $L$, resulting in poor quality of the enhanced image in over-exposed regions.

\subsection{Decomposition}
To address these limitations, we introduce the latent space decomposition strategy.
First, we apply a logarithmic transformation to input image, allowing Eq.~(\ref{eq1}) to be convert to:
\begin{equation}
S=\log (I)=\log (R) + \log (L)
\label{eq4}
\end{equation}
Through this transformation, $\log (R)$ and $\log (L)$ can be decomposed via an additive operation. 
Thus, when using convolution layers to predict the decomposed $\log (R)$ and $\log (L)$, the back-propagation coefficients for layers remain consistently 1.
This ensures the training procedure for the decomposition is more stable and efficient.
Considering the numerical explosion of conventional logarithmic transformations at extremely small values, namely low-light regions~(pixel values $\in[0,1]$), we employ a 1-pixel offset operation, as illustrated in Fig.~\ref{fig:first-img}(b).

Then, the decomposition process can be formulated as:
\begin{equation}
R, L = f_{dec}(S, p),\quad \tilde{S} = f_{rec}(R+L)
\label{eq5}
\end{equation}
where $f_{dec}(\cdot)$ denotes decomposition, and $f_{rec}(\cdot)$ means reconstruction.
Specifically, for an input image $I$, we first apply log transformation to obtain $S$.
An illumination prior $p\in \mathbb{R}^{H\times W\times 1}$ is then extracted as $p=Max_c(S)$, where $Max_c(\cdot)$ denotes the operation that calculates the maximum values of each pixel along channel dimension.
The features $S$ and $p$ are concatenated along the channel dimension and fused via a $3\times3$ convolutional layer, which projects the combined inputs into the latent space.
Two Transformer blocks subsequently decouple the resulting features into reflectance and illumination components.
Finally, a $3\times3$ convolutional layer reconstructs $R+L$ and maps the result back to the RGB space, producing the output $\tilde{S}$. Compared with decomposition in the RGB space, that in the latent space can leverage the strong non-linear modeling capability of deep neural networks, thus better adapting to complex lighting environments.

\begin{figure}[!t]
    \centering
    \includegraphics[width=1.0\linewidth]{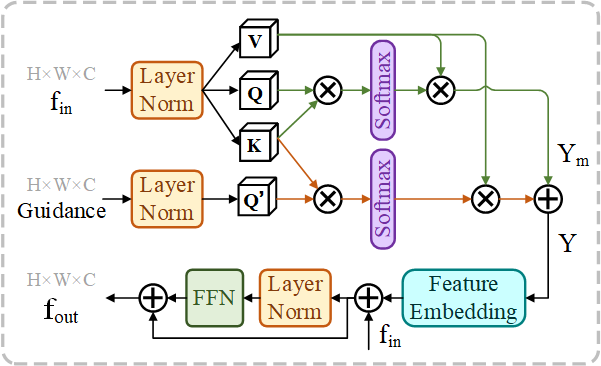}
    \caption{The architecture of the GFTB.}
    \label{gftb}
\end{figure}

\subsection{Enhancement}
Images captured under low-light conditions contain significant noise. We argue that this noise propagates to $R$ and $L$, so we revise Eq.~\ref{eq4} as:
\begin{equation}
\log(I)=\log(R+\hat{R})+\log(L+\hat{L}),
\label{eq6}
\end{equation}
where $\hat{R}$ and $\hat{L}$ are noise components in $R$ and $L$. 
Statistically, it is fair to hold the hypothesis that $\hat{R} \ll R$ and $\hat{L} \ll L$. 
Therefore, we can apply first-order Taylor expansion to extend the Eq.~(\ref{eq6}) as:
\begin{equation}
log(R+\hat{R}) \approx log(R)+\frac{\hat{R}}{R},
\label{eq7}
\end{equation}
\begin{equation}
log(L+\hat{L}) \approx log(L)+\frac{\hat{L}}{L}.
\label{eq8}
\end{equation}

Therefore, we design a Component Refiner~(CR) to enhance $\log (R)$ and $\log (L)$ by estimating noise components $\frac{\hat{R}}{R}$ and $\frac{\hat{L}}{L}$ in latent space, respectively. 
Note that the noise components are related to $R$ and $L$, it's necessary to provide a roughly estimated $R$ and $L$ as a guidance to estimate the corresponding noise component.
The guidance for $\frac{\hat{R}}{R}$ is given by $S_{mean}(i,j)=\frac{1}{3}\sum_{c\in(R,G,B)}S_c(i,j)$, while the guidance for $\frac{\hat{L}}{L}$ is provided by $S_{max}(i,j)=Max_{c\in(R,G,B)}(S_c(i,j))$.

As illustrated in Fig.~\ref{framework}(c), the component refiner~(CR) takes the estimated feature components and corresponding structural priors as dual inputs, and employs a three-scale U-shaped structure to enhance the input components. 
To fuse the prior features and component features, we further design the Guidance Fusion Transformer Block (GFTB), which is elaborated subsequently.

\textbf{Guidance Fusion Transformer Block.} 
As illustrated in Fig.~\ref{gftb}, we employ $S_{mean}$ and $S_{max}$ as guidance signals to preserve structural details in $R$ and maintain brightness distribution in $L$, respectively.
We utilize a 3$\times$3 convolutional layer to map the single-channel guidance signal to latent space, obtaining $G_L\in \mathbb{R}^{H\times W\times C}$ or $G_R\in \mathbb{R}^{H\times W\times C}$. 
Then, the guidance signal is incorporated into the decoupling components through the following process. 
Firstly, self-attention is applied to the input features to generate $Y_m$. 
We derive $Q, K, V$ by $Q=W^Qf_{in},K=W^Kf_{in},V=W^Vf_{in}$.
Simultaneously, the guidance signal is projected to a guidance query $Q'=W^GG$ for cross-attention.
The outputs of both attention paths are computed as:
\begin{equation}
Y_m=V\cdot\text{Softmax}(\frac{Q^TK}{\alpha_m}),
\end{equation}
\begin{equation}
Y_G=V\cdot\text{Softmax}(\frac{Q'^TK}{\alpha_G}),
\end{equation}
where $\alpha_m$ and $\alpha_G$ are learnable parameters.
The two outputs are combines as $Y = Y_m + Y_G$.
A feed-forward layer is then applied to $Y$, followed by the addition of positional encoding, to produce the final output feature $f_{out}$.

\subsection{Loss Function}
Our RGT model adopts a two-stage training paradigm, thus its total loss comprises two parts.

\textbf{Decomposition.} 
Following RetinexNet~\cite{RetinexNet}, the decomposition loss $L_{decom}$ consists of three terms, expressed as:
\begin{equation}
\mathcal{L}_{decom}=\mathcal{L}_{recon}+\lambda_1\mathcal{L}_{is}+\lambda_2\mathcal{L}_{iR}
\end{equation}
where $\lambda_1$ and $\lambda_2$ balance the contribution of each term.

The reconstruction loss $\mathcal{L}_{recon}$ enforces that the decomposed reflectance and illumination can accurately reconstruct the input image. 
The $\mathcal{L}_{is}$ ensures local-consistency and smoothness in the illumination map while while preserving strong structural details in the reflectance $R$. 
The $\mathcal{L}_{iR}$ promotes similarity between the reflectance estimates under different lighting conditions.
These terms are formulated as:
\begin{equation}
\begin{aligned}
\mathcal{L}_{recon}&=\sum_{i\in(l,n)}\|f_{rec}(R_i,L_i)-S_i\|_{1},  \\
\mathcal{L}_{is}&=\sum_{i\in(l,n)}\|\nabla L_i\circ exp(\alpha \nabla R_i)\|_{1}, \\
\mathcal{L}_{iR}&={\|R_{l}-R_{n}\|}_{1}
\end{aligned}
\end{equation}
where $\alpha$ represents a coefficient that balances the structure-awareness strength.

\textbf{Enhancement.} 
During the training of the enhancer, we employ L1 loss and perceptual loss $\mathcal{L}_p$\cite{Lploss} to guide the low-light enhancement process.
The enhancement loss $\mathcal{L}_{en}$ is defined as:
\begin{equation}
\mathcal{L}_{en}=\|S_{en}-S_{n}\|_{1}+\lambda_p \mathcal{L}_p(S_{en},S_{n})
\end{equation}

\begin{figure}[!t]
\centering
\includegraphics[width=1.0\linewidth]{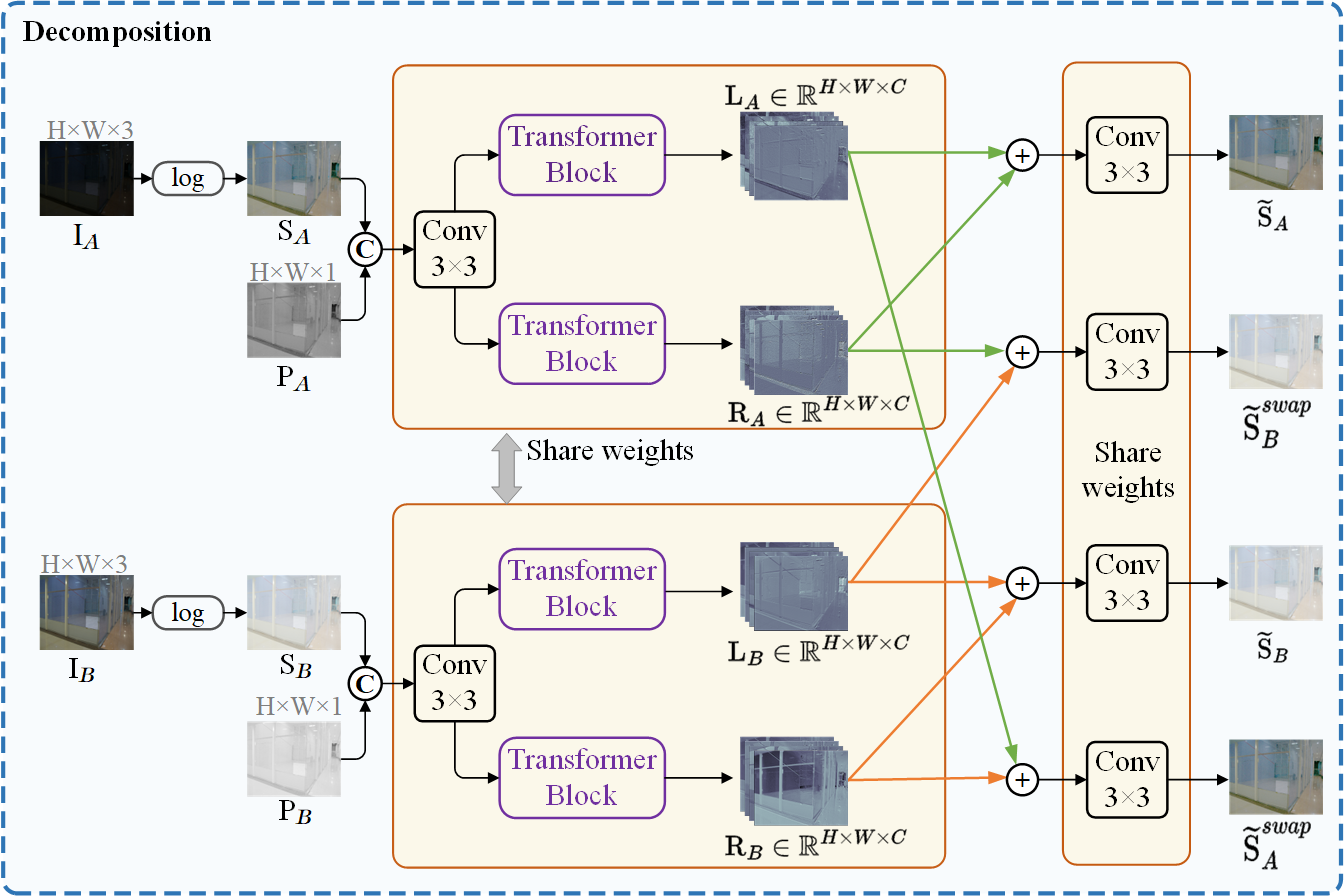}
\caption{Schematic illustration of illumination-reflectance component swapping between two content-identical images under different lighting conditions. Each input image is decomposed into reflectance ($R$) and illumination ($L$) components within the shallow spatial space. By exclusively exchanging the $L$ components between paired images, we can synthesize four novel recomposed results. Ideally, swapping only the $L$ components is sufficient to alter the brightness of the images.}
\label{fig7:decom}
\end{figure}

\begin{table*}[!t]
\centering
\caption{The PSNR of comparative decomposition on LOLv2 dataset. $f_{rec}$ represents the operation of reconstructing the image from $R$ and $L$ components. According to RetinexNet\cite{RetinexNet} requirements for the decomposed $R$ and $L$ components, the reconstructed images from $R$ and $L$ should match the input image corresponding to $L$.}
\scriptsize
\tabcolsep=0.15cm
    \begin{tabular}{l|cccc|cccc}
    \toprule
    \multirow{2}{*}{Methods} & \multicolumn{4}{c|}{LOLv2\_real} & \multicolumn{4}{c}{LOLv2\_syn} \\
     & $f_{rec}(R_l,L_l)$ & $f_{rec}(R_l,L_n)$ & $f_{rec}(R_n,L_l)$ & $f_{rec}(R_n,L_n)$ & $f_{rec}(R_l,L_l)$    & $f_{rec}(R_l,L_n)$ & $f_{rec}(R_n,L_l)$ & $f_{rec}(R_n,L_n)$  \\
    \midrule
    RetinexNet\cite{RetinexNet} & \underline{38.56} & \underline{23.90} & \underline{35.47} & \underline{29.50} & \underline{36.44} & \underline{20.37} & \underline{26.59} & \underline{30.28} \\
    RetinexDIP\cite{retinexdip} & 16.61 & 14.91 & 14.49 & 18.05 & 23.75 & 14.43 & 23.52 & 20.47 \\
    URetinexNet\cite{uretinexnet} & 34.85 & 19.40 & 31.47 & 24.76 & 26.78 & 17.39 & 25.36 & 22.89 \\\hdashline
    & & & & & & & & \\[-0.8em]
    RGT & \textbf{69.45} & \textbf{54.77} & \textbf{68.34} & \textbf{59.05} & \textbf{58.22} & \textbf{51.11} & \textbf{56.82} & \textbf{53.72} \\
    \bottomrule
    \end{tabular}%
  \label{tab:decom}%
\end{table*}%
\begin{figure*}[!t]
\centering
\includegraphics[width=1.0\linewidth]{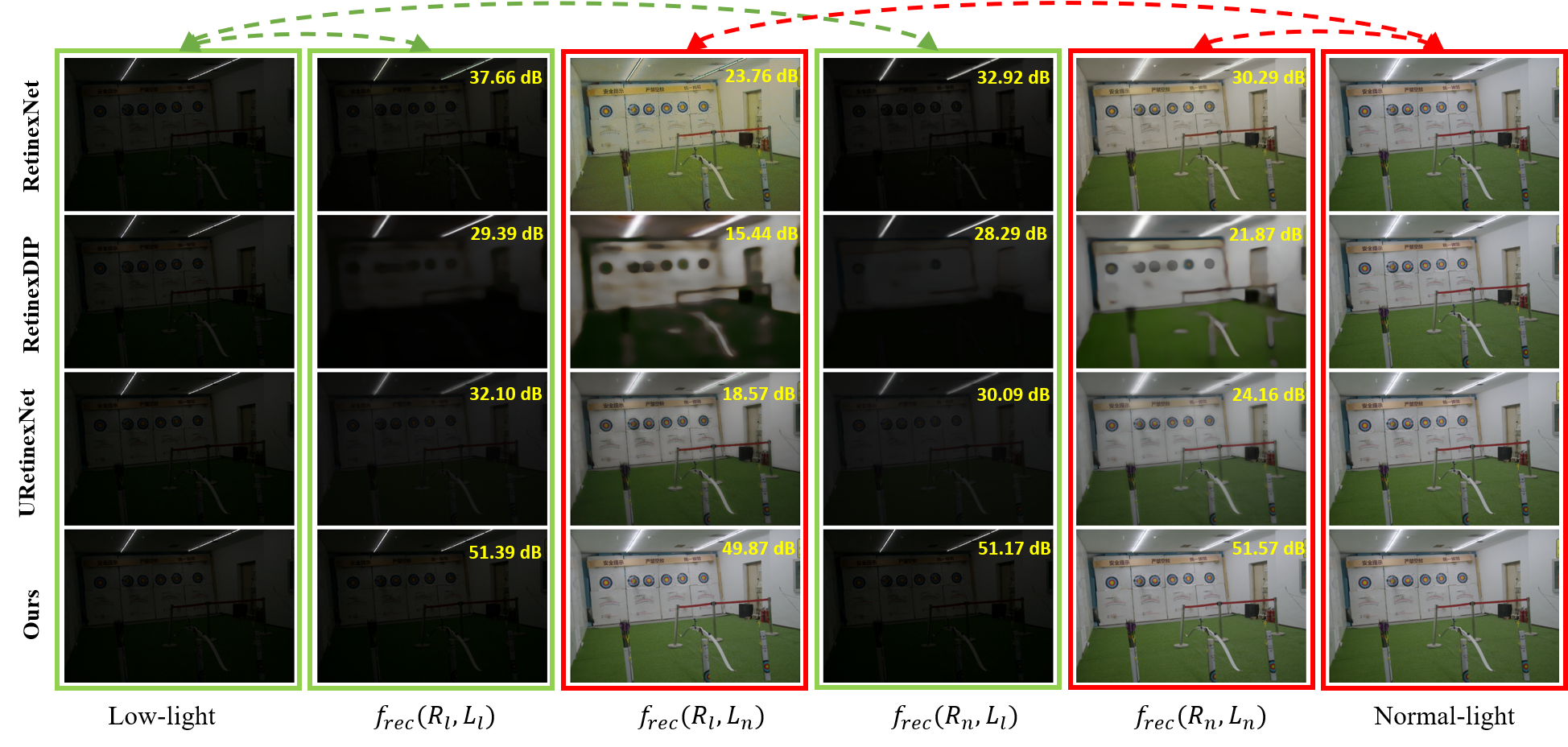}
\caption{Visual comparisons of the decomposed results by different methods on LOLv2\_real. The images within the green boxes should be same and the images within the red boxes should be same.}
\label{fig:decom}
\end{figure*}

\section{Experiment}\label{Experiments and Results}
\subsection{Datasets and Evaluation Metrics}\label{Datasets and Evaluation Metrics}

\textbf{Datasets.}
We evaluate our method on four widely used paired low-light image datasets, including LOLv2\cite{LOLv2}, HDR+(480p)\cite{hdr}, SDSD-indoor\cite{sdsd} and SICE\cite{SICE}.
We also conduct evaluation on unpaired real-world datasets, including DICM\cite{DICM}, LIME\cite{LIME}, MEF\cite{MEF}, NPE\cite{NPE}, and VV\cite{VV}.\\
\textbf{LOLv2} contains two subsets, Real and Synthetic. The training-to-test data pair ratios are set as 689:100 for LOLv2\_real and 900:100 for LOLv2\_syn.\\
\textbf{HDR+(480p)} provides ISP-untreated inputs that simulate low-light conditions, making it suitable for low-light enhancement evalution.
The dataset includes 675 training pairs and 247 testing pairs.\\
\textbf{SDSD-indoor} uses a static(indoor) subset, with 62 image pairs for training and 6 for testing.\\
\textbf{SICE} Follows FECNet\cite{FECNet}, where the medium-exposure image is taken as the ground truth. 
The shortest-exposure image is used as the low-light input. The dataset contains 452 training pairs and 50 testing pairs.

\textbf{Evaluation metrics.}
For paired datasets, we adopt Peak Signal-to-Noise Ratio (PSNR) and Structural Similarity (SSIM)~\cite{ssim} as evalution metrics. The Learned Perceptual Image Patch Similarity (LPIPS) metric, based on AlexNet~\cite{alexnet}, is also employed to quantify perceptual similarity.
For unpaired datasets, we utilize CLIP-IQA~\cite{clip-iqa} and NIQE~\cite{niqe} to evaluate the quality of the enhanced images, as these metrics do not require paired ground-truth references.

\subsection{Implementation Details}\label{Model Settings and Training Details}
Our method is implemented by PyTorch and trained on an NVIDIA 4090 GPU with the Adam optimizer ($\beta_1 = 0.9$ and $\beta_2 = 0.999$) for $1.5 \times 10^5$ iterations.
The initial learning rate is $2\times10^{-4}$, which is decayed to $1\times10^{-6}$ via cosine annealing. 
Training samples are (256$\times$256) patches randomly cropped from low-/normal-light image pairs, with a batch size of 4.
Data augmentation includes random rotation and flipping.
The two-stage training protocol proceeds as follows: (i) Decomposition phase: train the decomposition network with loss weights $\lambda_1=0.1$ and $\lambda_2=1$, outputting $R$ and $L$ components with channel dimension $C=40$. 
(ii) Enhancement phase: the decomposition network parameters frozen, the $CR$ is trained with $\lambda_p=0.01$.

\subsection{Comparison with SOTA Models}\label{Comparison with SOTA Models}
We perform extensive comparisons between our method and SOTA low-light enhancement algorithms, covering both quantitative and visual evaluations across paired and unpaired datasets.

\textbf{Decomposition quality comparison.}
To validate the effectiveness of our latent space decomposition strategy, we conduct paired-image experiments through illumination~($L$) component swapping. 
According to Retinex theory, swapping only the $L$ components should alter solely the image brightness without affecting their content information. 
As shown in Fig.~\ref{fig7:decom}, given paired images $I_A$ and $I_B$, our latent space decomposition strategy yields illumination-reflectance component pairs $\{L_A,R_A\}$and $\{L_B,R_B\}$. 
By exclusively exchanging their illumination components ($L_A\leftrightarrow L_B$), we generate four recomposed variants through the following cross-synthesis protocol:
\begin{equation}
\begin{aligned}
f_{rec}(R_l,L_l)&=\tilde{S}_{l},\\
f_{rec}(R_l,L_n)&=\tilde{S}_{l}^{swap},\\
f_{rec}(R_n,L_n)&=\tilde{S}_{n},\\
f_{rec}(R_n,L_l)&=\tilde{S}_{n}^{swap},\\
\end{aligned}
\end{equation}Where $f_{rec}$ represents the reconstruction operation, and superscript $swap$ represents recomposed outputs generated through $L$ component exchange. 
We emphasize that under ideal conditions, if $\tilde{S}_{l}(\text{or}~\tilde{S}_{n})$ and $\tilde{S}_{l}^{swap}(\text{or}~\tilde{S}_{n}^{swap})$ are consistent, it indicates that the $L$ component dominates the brightness characteristics of the image, while the $R$ component determines the content information, thus validating the accuracy of the image decoupling process.
As shown in Table~\ref{tab:decom}, our decomposition strategy achieves the best performance.
We also compare decomposition quality visually with three representative approaches. 
As shown in Fig.~\ref{fig:decom}, competing methods exhibit inferior decomposition performance when $R$ and $L$ components are examined independently, with issues including exposure inconsistency and structural degradation. In contrast, our approach leverages latent-space additive coupling to enable stable training and precise decomposition of $R$ and $L$. 
This robust decomposition lays a solid foundation for high-quality enhancement, particularly in handling overexposed regions.
More comprehensive experimental analysis will be conducted in the ablation experiment section.

\begin{table*}[!t]
\caption{Quantitative comparisons PSNR/SSIM on LOLv2\cite{LOLv2}, HDR+\cite{hdr}, SDSD(indoor)\cite{sdsd} and SICE\cite{SICE} datasets. The highest result is in \textbf{bold} while the second highest result is \underline{underlined}.} 
\label{Quantitative}
\tabcolsep=0.15cm
\centering
\begin{tabular}{lccc *{5}{cc}}
\toprule
\multirow{2}{*}{Methods} & 
\multicolumn{2}{c}{LOLv2\_real} & 
\multicolumn{2}{c}{LOLv2\_syn} & 
\multicolumn{2}{c}{HDR+(480p)} & 
\multicolumn{2}{c}{SDSD-indoor} & 
\multicolumn{2}{c}{SICE} &
\multicolumn{3}{c}{Complexity} \\
\cmidrule(lr){2-3} \cmidrule(lr){4-5} \cmidrule(lr){6-7} \cmidrule(lr){8-9} \cmidrule(lr){10-11} \cmidrule(lr){12-14}
    & PSNR & SSIM  & PSNR & SSIM & PSNR & SSIM & PSNR & SSIM  & PSNR & SSIM & FLOPs(G) & Params(M) & Time(ms) \\
    \midrule
    RetinexNet\cite{RetinexNet} & 16.52 & 0.674 & 19.11 & 0.805 & 19.54 & 0.678 & 25.88 & 0.832 & 19.71 & 0.686 & 587.47 & 0.84 & 3.36 \\
    LLFlow\cite{llflow} & 20.41 & 0.829 & 23.14 & 0.930 & 22.96 & 0.762 & 22.10 & 0.801 & 19.99 & 0.685 & 358.40 & 17.42 & 122.70\\
    MirNetv2\cite{MIRNetv2} & 22.00 & 0.818 & 24.86 & 0.925 & 22.87 & 0.765 & 29.68 & 0.870 & 24.29 & 0.767 & 140.21 & 5.86 & 61.69\\
    SNR-Net\cite{snrnet} & 21.48 & 0.849 & 24.14 & 0.928 & 22.50 & 0.764 & 29.44 & 0.894 & 21.78 & 0.729 & 26.35 & 4.01 & 6.25 \\
    GSAD\cite{gasd} & 22.20 & 0.811 & 22.46 & 0.897 & 18.03 & 0.781 & -- & -- & 15.11 & 0.553 & 134.04 & 18.42 & 164.99 \\
    LLFormer\cite{llformer} & 21.86 & 0.816 & 25.32 & 0.933 & 23.38 & 0.768 & 29.18 & 0.867 & 23.44 & 0.739 & 22.52 & 24.55 & 70.28\\
    Cotf\cite{cotf} & 19.25 & 0.774 & 22.91 & 0.892 & 22.31 & 0.737 & 28.41 & 0.875 & 22.50 & 0.769 & 1.81 & 0.31 & 3.56 \\
    CSPN\cite{CSPN} & 21.59 & 0.859 & 25.34 & 0.935 & 25.18 & 0.906 & \underline{30.89} & \underline{0.895} & 23.53 & 0.775 & 60.92 & 1.40 & 118.41\\
    AttnFlow\cite{AttnFlow} & 20.23 & \underline{0.877} & 24.28 & \textbf{0.954} & 25.75 & 0.906 & 25.83 & 0.866 & 21.08 & 0.717 & 286.37 & 17.42 & 624.55\\
    IGDFormer\cite{IGDFormer} & 22.73 & 0.833 & 25.33 & 0.937 & 24.61 & 0.897 & 28.88 & 0.838 & 23.28 & 0.756 & 38.23 & 3.55 & 77.85\\
    RetinexMamba\cite{retinexmamba} & 22.70 & 0.858 & \underline{25.79} & 0.933 & 24.92 & 0.899 & 29.63 & 0.878 & 24.64 & 0.768 & 45.98 & 4.59 & 164.88\\
    CIDNet-op\cite{cidnet} & 23.11 & 0.849 & 25.44 & 0.935 & 21.17 & 0.866 & 21.69 & 0.877 & 23.67 & 0.751 & 7.57 & 1.88 & 18.37\\
    CIDNet-wp\cite{cidnet} & \textbf{23.81} & 0.865 & 25.28 & 0.934 & -- & -- & 22.30 & 0.861 & 24.19 & \underline{0.772} & 7.57 & 1.88 & 18.37\\
    RetinexFormer\cite{retinexformer} & 22.80 & 0.840 & 25.67 & 0.939 & \underline{26.18} & \underline{0.910} & 29.77 & \underline{0.895} & \underline{24.84} & 0.765 & 15.57 & 1.61 & 18.96\\
    \hdashline
    RGT & \underline{23.80} & \textbf{0.896} & \textbf{26.18} & \underline{0.941} & \textbf{26.97} & \textbf{0.924} & \textbf{31.31} & \textbf{0.896} & \textbf{26.40} & \textbf{0.827} & 50.30 & 4.79 & 55.71\\
    \bottomrule
  \end{tabular}
  \normalsize
\end{table*}

\begin{figure*}[!t]
    \centering
    \includegraphics[width=\textwidth]{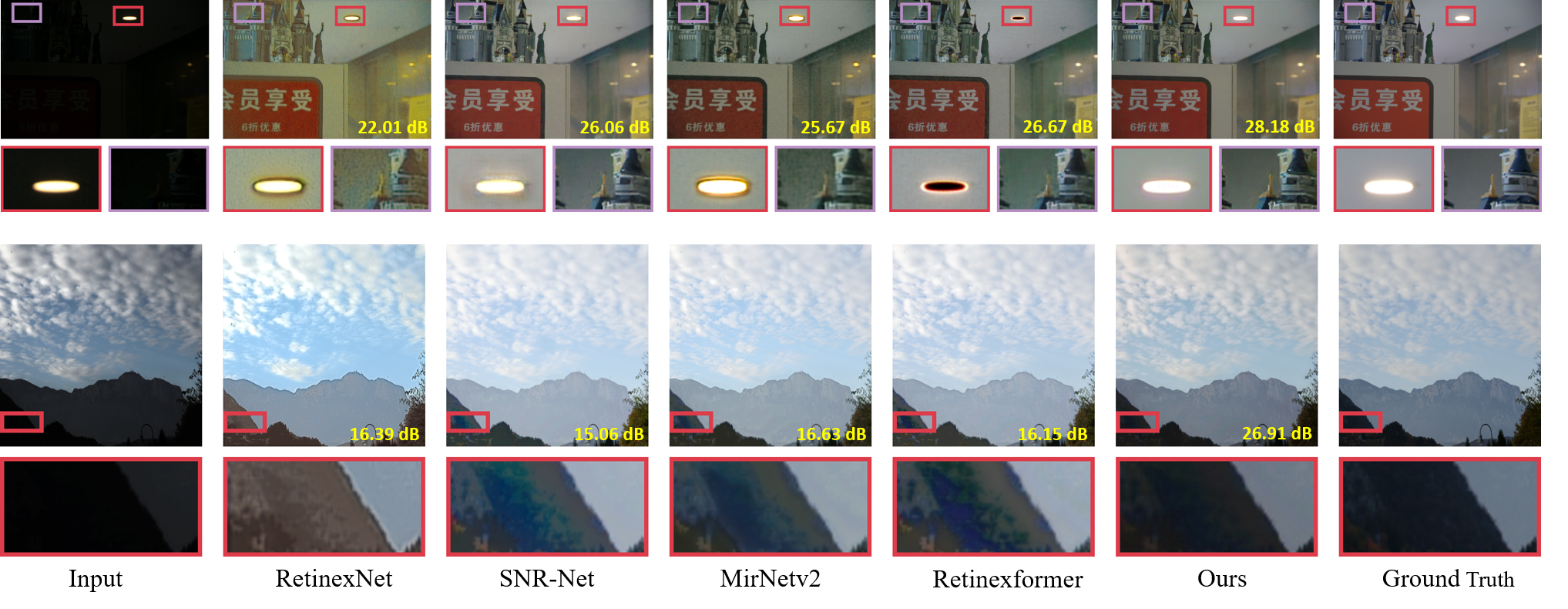}
    \caption{Visual comparisons of the enhanced results by different methods on LOLv2\_real~(top) and LOLv2\_syn~(bottom).}
    \label{LOLv2}
\end{figure*}

\begin{figure*}[!t]
    \centering
    \includegraphics[width=0.95\linewidth]{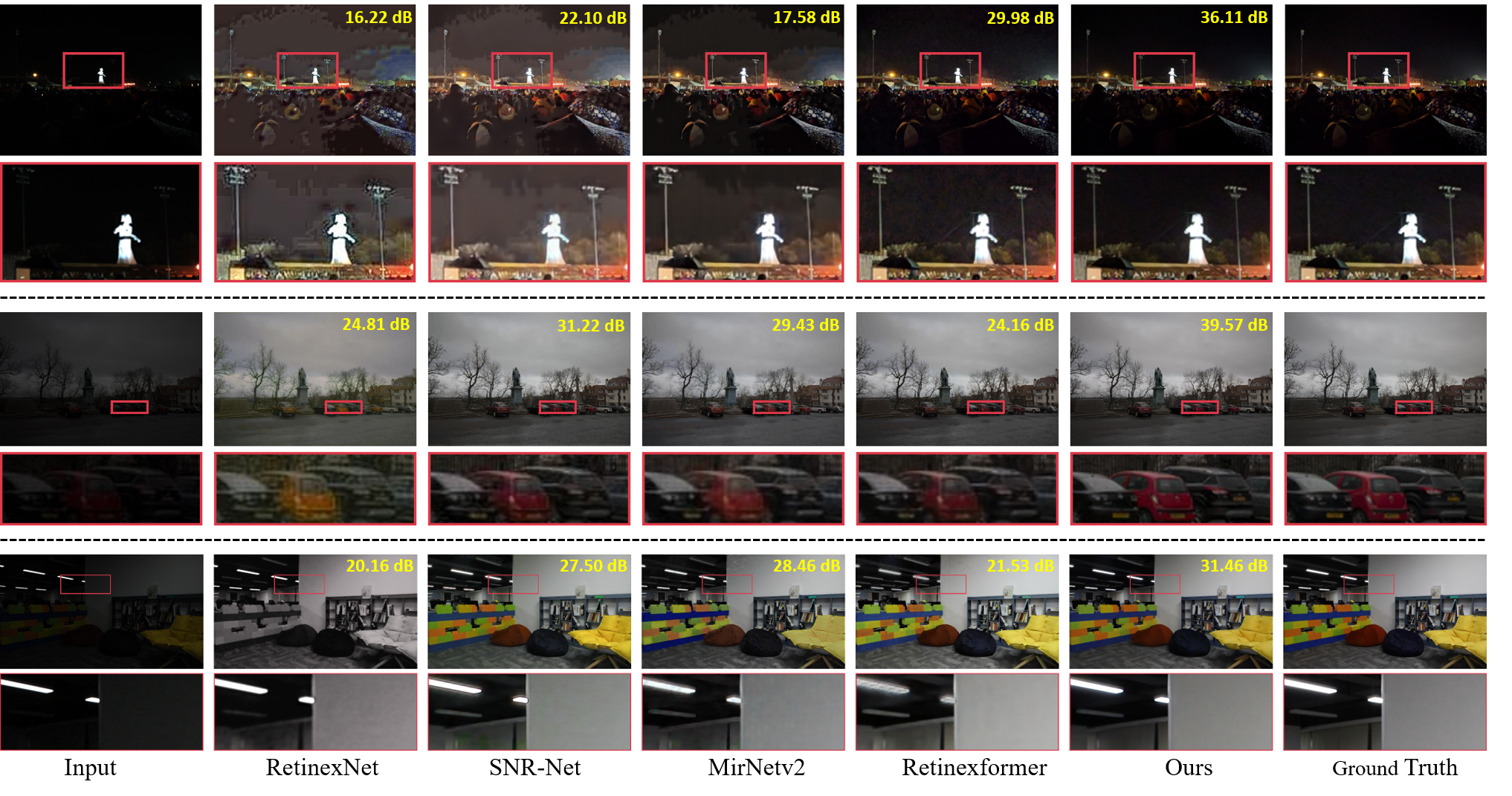}
    \caption{Visual comparisons of the enhanced results by different methods on HDR+(480p)~(top), SICE~(middle) and SDSD-indoor~(bottom)}
    \label{hdr_sice}
\end{figure*}

\begin{table}[!t]
    \caption{Quantitative comparisons LPIPS$\downarrow$ on LOLv2~\cite{LOLv2}, HDR+(480p)~\cite{hdr}, SDSD(indoor)~\cite{sdsd} and SICE~\cite{SICE} datasets. The highest result is in \textbf{bold} while the second highest result is \underline{underlined}.}
    \centering
    \scriptsize
    \tabcolsep=0.1cm
    \begin{tabular}{c|ccccc}
    \hline
    & &&&& \\[-0.8em]
    Methods & LOLv2\_real & LOLv2\_syn & HDR+(480p) & SDSD-indoor & SICE  \\
    & &&&& \\[-0.8em]
    \hline
    & &&&& \\[-0.8em]
    SNR-Net\cite{snrnet} & 0.122  & 0.061  & 0.238  & 0.230  & \underline{0.206}  \\
    CSPN\cite{CSPN} & 0.156  & 0.055  & 0.071  & 0.181  & 0.235  \\
    IGDFormer\cite{IGDFormer} & 0.158  & 0.074  & 0.086  & 0.227  & 0.237  \\
    AttnFlow\cite{AttnFlow} & 0.146  & \underline{0.050}  & \underline{0.064}  & 0.223  & 0.298  \\
    RetinexMamba\cite{retinexmamba} & 0.147 & 0.062 & 0.073 & 0.184 & 0.230 \\
    CIDNet\cite{cidnet} & 0.131 & 0.055 & 0.086 & 0.198 & 0.258 \\
    RetinexFormer\cite{retinexformer} & 0.169 & 0.064 & 0.069 & \textbf{0.116} & 0.229 \\
    \hdashline
    RGT & \textbf{0.102} & \textbf{0.046} & \textbf{0.047} & \underline{0.119} & \textbf{0.127} \\ 
    \hline
    \end{tabular}
    \label{tab3:lpips}
\end{table}

\begin{table*}[!t]
    \caption{Quantitative comparisons CLIP-IQA$\uparrow$/NIQE$\downarrow$ on DICM~\cite{DICM}, LIME~\cite{LIME}, MEF~\cite{MEF}, NPE~\cite{NPE} and VV~\cite{VV} datasets. The highest result is in \textbf{bold} while the second highest result is \underline{underlined}.}
    \centering
    \scriptsize
    \tabcolsep=0.2cm
    \begin{tabular}{c|cc|cc|cc|cc|cc}
    \hline
    &  &  &  &  &  &  &  &  &  &\\[-0.8em]
    \multirow{2}{*}{Methods} & \multicolumn{2}{c|}{DICM} & \multicolumn{2}{c|}{LIME} & \multicolumn{2}{c|}{MEF} & \multicolumn{2}{c|}{NPE} & \multicolumn{2}{c}{VV} \\ 
    & CLIP-IQA$\uparrow$ & NIQE$\downarrow$ & CLIP-IQA$\uparrow$ & NIQE$\downarrow$ & CLIP-IQA$\uparrow$ & NIQE$\downarrow$ & CLIP-IQA$\uparrow$ & NIQE$\downarrow$ & CLIP-IQA$\uparrow$ & NIQE$\downarrow$ \\ 
    &  &  &  &  &  &  &  &  &  &\\[-0.8em]
    \hline
    &  &  &  &  &  &  &  &  &  &\\[-0.8em]
    SNR-Net\cite{snrnet} & 0.599 & 4.71 & 0.665 & 5.74 & 0.633 & 4.18 & 0.544 & 4.32 & 0.304 & 9.87 \\
    CSPN\cite{CSPN} & 0.659 & 4.10 & 0.626 & 4.12 & 0.604 & 4.42 & 0.668 & 3.97 & 0.297 & 3.24 \\
    IGDFormer\cite{IGDFormer} & 0.652  & \textbf{3.77}  & 0.665  & \underline{3.96}  & 0.688  & 3.51  & 0.530  & 4.30  & 0.313  & 2.99  \\
    AttnFlow\cite{AttnFlow} & 0.730  & 4.54  & \underline{0.712}  & 4.46  & 0.746  & 4.64  & 0.683  & 4.24  & 0.352  & 3.00  \\
    RetinexMamba\cite{retinexmamba} & 0.665  & 3.79  & 0.691  & \underline{3.96}  & 0.673  & \textbf{3.35}  & 0.573  & 3.92  & \textbf{0.384}  & \underline{2.90}  \\
    CIDNet\cite{cidnet} & \textbf{0.743}  & 3.99  & 0.675  & 4.16  & 0.664  & 3.72  & \underline{0.758}  & \underline{3.73}  & 0.299  & 3.19  \\
    RetinexFormer\cite{retinexformer} & 0.711  & 3.79  & 0.687  & 4.13  & \underline{0.776}  & 3.56  & 0.691  & 3.74  & 0.297  & 3.21  \\
    &  &  &  &  &  &  &  &  &  &\\[-0.8em]
    \hdashline
    &  &  &  &  &  &  &  &  &  &\\[-0.8em]
    RGT & \underline{0.732}  & \underline{3.78}  & \textbf{0.723}  & \textbf{3.85}  & \textbf{0.810}  & \underline{3.42}  & \textbf{0.791}  & \textbf{3.72}  & \underline{0.361}  & \textbf{2.87}  \\
    &  &  &  &  &  &  &  &  &  &\\[-0.8em]
    \hline
    \end{tabular}
    \label{tab:unpaired-niqe-clipiqa}
\end{table*}

\begin{figure*}[!t]
\centering
\includegraphics[width=0.95\linewidth]{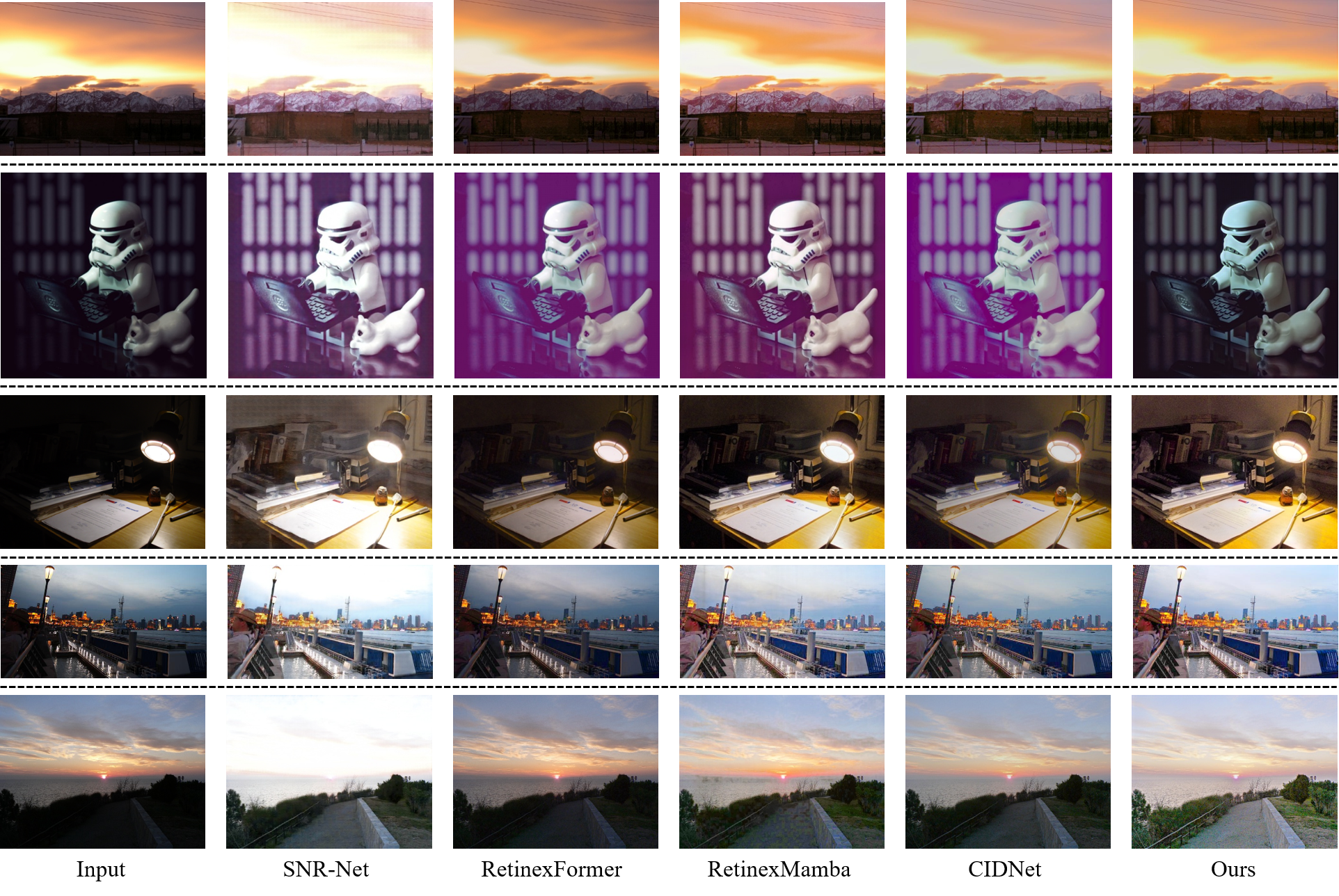}
\caption{Visual comparisons of the enhanced results by different methods on unpaired dataset (DICM, LIME, MEF, NPE and VV).}
\label{fig:unpaired}
\end{figure*}

\begin{figure*}[t]
    \centering
    \includegraphics[width=0.95\linewidth]{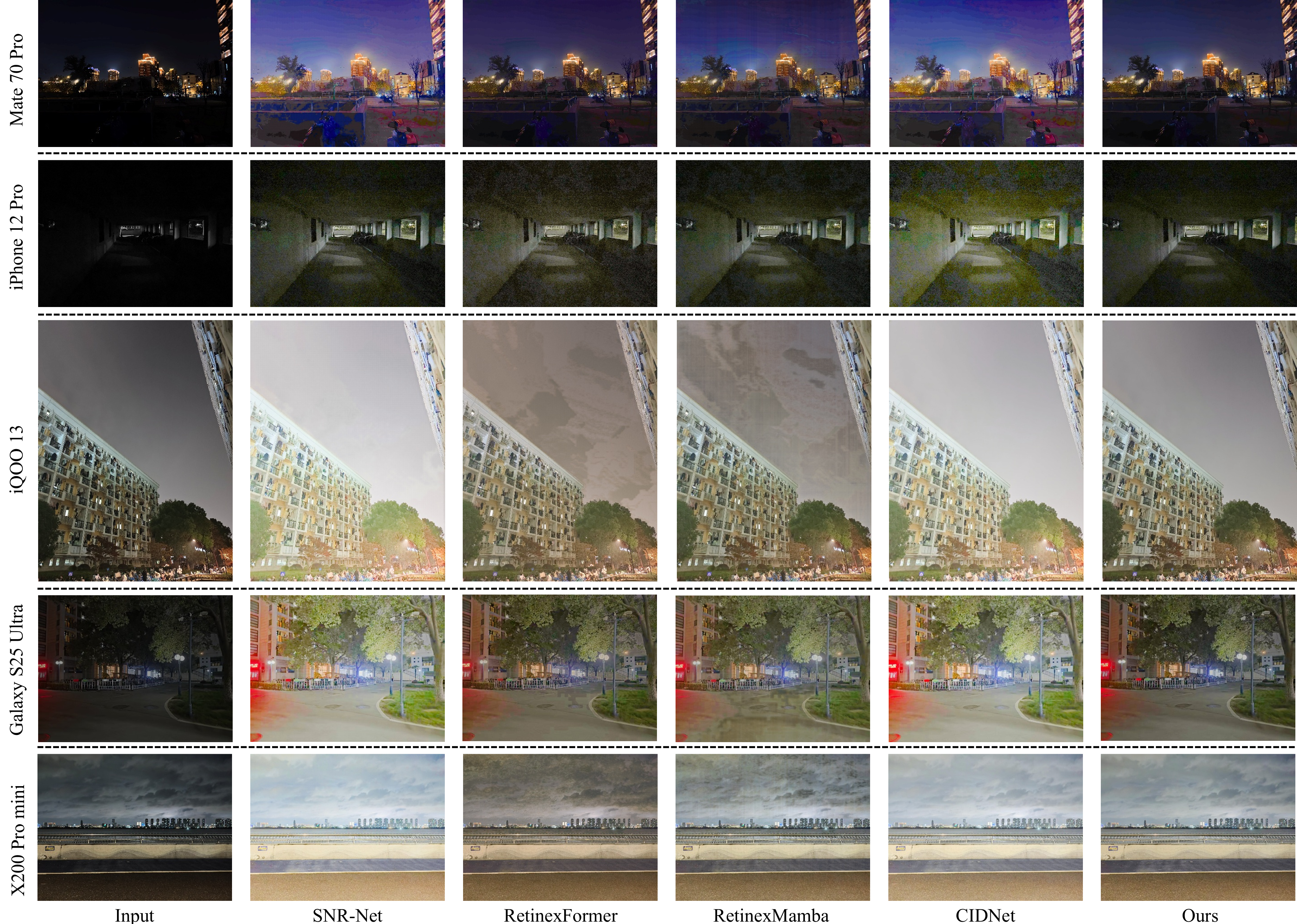}
    \caption{Visual comparison of various methods on real-world images from various mobile devices.}
    \label{fig:real-phone}
\end{figure*}

\textbf{Comparison on paired datasets.}
As shown in Table~\ref{Quantitative}, the high-quality decomposition also leads to outstanding PSNR and SSIM performance in low-light image enhancement task, outperforming all competing methods on four benchmark datasets while securing second-best results on the remaining one. 
On the SICE dataset, where shortest-exposure images serve as low-light inputs, our method achieves a 1.36 dB PSNR improvement compared with the second-ranked method. 
This gain verifies that latent-space decomposition allows $R$ and $L$ components to break free from numerical constraints effectively. 
The nonlinear fusion capability of the convolutional neural network in our model further enhances adaptability to extremely low-light environments.
In terms of perceptual quality, the RGT model achieves the best LPIPS results across all five datasets, indicating that the enhanced images are better aligned with human perception~(Table~\ref{tab3:lpips}).

Visual comparisons between our method and SOTA approaches are conducted on five benchmark datasets.
As shown in Fig.~\ref{LOLv2} and Fig.~\ref{hdr_sice}, existing methods face difficulties in effectively enhancing extremely dark regions effectively while suppressing noise, often resulting in color distortion or insufficient enhancement. 
For instance, Retinexformer relies on a multiplicative  relationship to derive $R$ and $L$ in RGB space, failing to decompose images into accurate $R$ and $L$ components. This deficiency leads to over-enhancement and noisy artifacts. 
Moreover, previous methods apply clipping operations to $R$ and $L$, which frequently causes artifacts in overexposed regions during enhancement.

\textbf{Comparison on unpaired datasets.}
For unpaired datasets~(DICM, LIME, MEF, NPE and VV), ground-truth images are unavailable. 
To evaluate performance under realistic conditions, images are enhanced using models trained on LOLv2 or HDR+(480p), and assess the results with the CLIP-IQA and NIQE metrics. 
As reported in Table~\ref{tab:unpaired-niqe-clipiqa}, the proposed method consistently ranks among the top two across all metrics, demonstrating that it produces more natural-looking results with superior overall visual quality. 
Representative images from each dataset are selected for visual comparison. 
As illustrated in Fig.~\ref{fig:unpaired}, the advantages of latent-space decomposition allow the proposed method to cope robustly with very dark and challenging lighting, achieving balanced global enhancement without over-exposure.

\textbf{Comparison on real-world data.}
Furthermore, we collected real low-light images using five mobile devices and selected five methods for comparison, with the results presented in Fig.~\ref{fig:real-phone}.
Visual inspections reveal that in comparison with competing methods such as RetinexFormer and RetinexMamba, the enhanced images generated by the RGT model exhibit fewer artifacts and lower noise levels. This observation demonstrates the strong generalization capability of our proposed method.

\begin{table}[!t]
\renewcommand{\arraystretch}{1.20}
  \centering
  \caption{Ablation study on decomposition strategies using the LOLv2\_real benchmark. Subscripts $l/n$ denote components derived from low-light and normal images respectively.}
\resizebox{1.0\linewidth}{!}{
    \begin{tabular}{c|cccc|cc}
    \hline
     Methods & \multicolumn{1}{c}{$f_{rec}(R_l,L_l)$} & \multicolumn{1}{c}{$f_{rec}(R_l,L_n)$} & \multicolumn{1}{c}{$f_{rec}(R_n,L_l)$} & \multicolumn{1}{c|}{$f_{rec}(R_n,L_n)$} & PSNR & SSIM\\
    \hline
     (0) & 40.35 & 30.84 & 39.91 & 38.35 & 21.76 & \underline{0.844} \\
     (1) & 50.16 & 23.56 & 42.76 & 38.52 & 22.30 & \underline{0.844}  \\
     (2) & \underline{52.73} & \underline{48.19} & \underline{51.77} & \underline{51.58} & \underline{23.27} & 0.811  \\
     (3) & 51.12 & 22.83 & 39.03 & 50.53 & 22.63 & 0.832  \\ \hdashline
     Ours & \textbf{69.45} & \textbf{54.77} & \textbf{68.34} & \textbf{59.05} & \textbf{23.80} & \textbf{0.896}  \\
    \hline
    \end{tabular}%
    }
  \label{ab:decom2}%
\end{table}%

\subsection{Ablation Study}
We conducted ablation studies on the LOLv2\_real dataset to validate the effectiveness of each component of RGT.

\begin{figure*}[!t]
    \centering
    \includegraphics[width=0.95\linewidth]{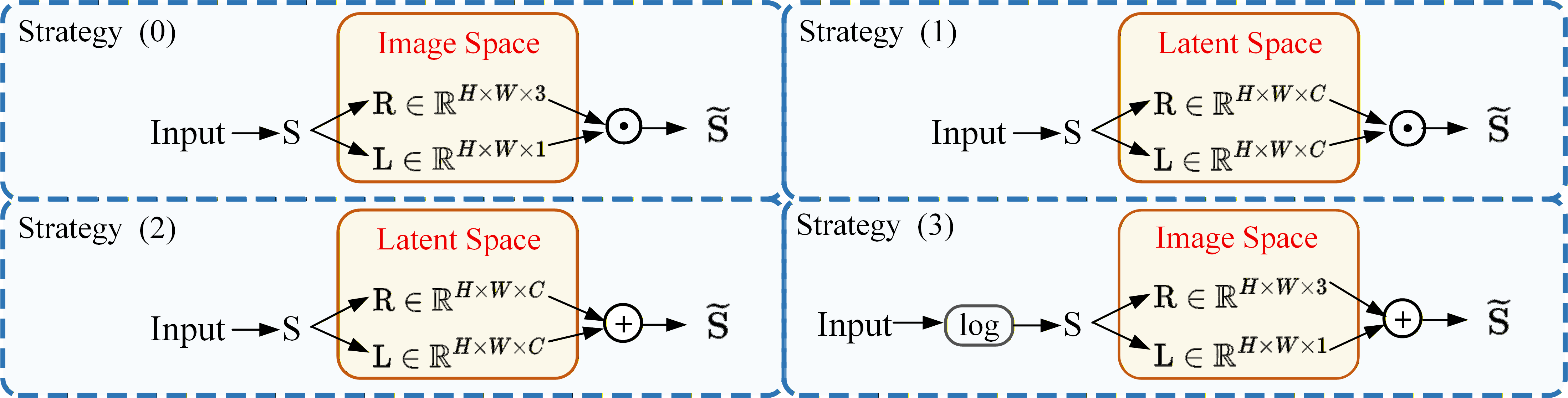}
    \caption{The overview of four variant decomposition strategies.}
    \label{fig:strategy}
\end{figure*}

\begin{table*}[!t]
\renewcommand{\arraystretch}{1.00}
\centering
\caption{Ablation study on decomposition strategies using the Kalantri dataset. Based on the types of input image pairs, the experiments can be divided into three groups:(i) Low$\leftrightarrow$Middle; (ii) Middle$\leftrightarrow$High; (iii) Low$\leftrightarrow$High. Each group employs three decomposition strategies for the experiments, including the classical Retinex strategy (variant~(1)), variant~(2), and the full strategy~(Ours).}
\resizebox{0.9\linewidth}{!}{
\begin{tabular}{c|c|cccc|c}
\hline
Inputs    & Methods & $f_{rec}(R_l,L_l)$ &$f_{rec}(R_l,L_n)$ &$f_{rec}(R_n,L_l)$ & $f_{rec}(R_n,L_n)$  &Avg     \\ \hline
\multirow{3}{*}{Low$\leftrightarrow$Middle} & (1)  & 57.27    & 46.19     & 46.18    & 57.33       & 51.74       \\
& (2)    & \underline{64.74}    & \underline{59.52}          & \underline{62.16}          & \underline{61.03}          & \underline{61.86}          \\
& Ours       & \textbf{68.00} & \textbf{61.92} & \textbf{66.86} & \textbf{62.73} & \textbf{64.88} \\ \hdashline
\multirow{3}{*}{Middle$\leftrightarrow$High} & (1)       & 58.38     & 41.27     & 39.42      & 50.39    & 47.37     \\
& (2)    & \underline{61.03}    & \underline{57.39}          & \underline{58.29}          & \underline{59.18}          & \underline{58.97}          \\
& Ours       & \textbf{62.73} & \textbf{57.62} & \textbf{61.38} & \textbf{58.78} & \textbf{60.13} \\ \hdashline
\multirow{3}{*}{Low$\leftrightarrow$High}   & (1)    & 57.27  & 42.08   & 37.71   & 50.40   & 45.87    \\
& (2)    & \underline{64.74}       & \underline{58.21}     & 
\underline{59.27}     &
\underline{58.19}     & 
\underline{60.10}      \\
& Ours     & \textbf{68.00} & \textbf{57.09} & \textbf{65.47} & \textbf{58.78} & \textbf{62.34} \\ \hline
\end{tabular}
}
\label{ab:hdr}%
\end{table*}

\textbf{Effectiveness of the decomposition strategy.} 
To validate the efficacy of our latent space decomposition strategy, we design four targeted decoupling variants, as detailed in Fig.~\ref{fig:strategy}.
\begin{itemize}
\item (0) means standard Retinex decomposition strategy: Images are decoupled in the pixel domain into a 3-channel reflectance~($R$) component and 1-channel illumination~($L$) component, with reconstruction through multiplication;
\item (1) denotes that we decouple image by multiplication operation in latent space; 
\item (2) means that we decouple image by additive operation in latent space without log transform on image;
\item (3) denotes that we decouple image by additive operation in RGB space with log transform.
\end{itemize}

As shown in Table~\ref{ab:decom2}, the multiplicative operation degrades both the decomposition quality and the final enhancement performance, owing to ineffective gradient back-propagation. 
Although variant (3) yields acceptable results in decomposition and enhancement, it cannot fully unleash the model’s potential as it lacks the necessary logarithmic transformation.
Decomposition in the image space further struggles with bright regions under linear reconstruction constraints, such as $f_{rec}(R_l,L_n)$ only get 22.83 dB.
By contrast, our proposed method achieves superior performance in both decomposition and enhancement, benefiting from improved gradient flow and the ability to handle highlight regions effectively.

Low-light enhancement datasets typically contain paired low-light and normal-light images, thus failing to validate the applicability of image decomposition strategies to over-exposed images—another common extreme illumination scenario. 
To bridge this gap, we conduct supplementary experiments on the Kalantri dataset~\cite{kalantari2017deep}, 
containing 74 static scenes with low-normal-high triplet images. 
To fully validate the effectiveness of our proposed latent space decomposition strategy, we design three image combinations based on brightness differences: (i) Low$\leftrightarrow$Middle; (ii) Middle$\leftrightarrow$High; (iii) Low$\leftrightarrow$High.
For each combination, we conduct experiments with three decomposition strategies, including the classical Retinex strategy (\textit{i.e.}, variant~(1)), variant~(2), and the full strategy. 
Notably, the model use frozen weights from Table~\ref{ab:decom2}, without retraining or fine-tuning. 
As shown in Table~\ref{ab:hdr}, the proposed latent decomposition strategy achieves the best quantitative results across three combinations, outperforming the classical Retinex decomposition strategy by at least 10dB. 
This sufficiently demonstrates that our method is applicable to images with varying brightness and exhibits excellent generalization ability.
Compared with the classic Retinex image decoupling strategy, our method yields more accurate $R$ and $L$ components.
Furthermore, swapping the $L$ component between two images of different brightness enables precise adjustment of image brightness while preserving richer details.

\begin{table}[!t]
\renewcommand{\arraystretch}{1.20}
  \centering
  \caption{Ablation study on decomposition strategies using the LOLv2\_real benchmark. Subscripts $l/n$ denote components derived from low-light and normal images respectively. * denotes the model employs the latent space decomposition strategy.}
  \resizebox{1.0\linewidth}{!}{
    \begin{tabular}{c|cccc|c}
    \hline
    Methods & $f_{rec}(R_l,L_l)$ &$f_{rec}(R_l,L_n)$ &$f_{rec}(R_n,L_l)$ & $f_{rec}(R_n,L_n)$  &Avg\\
    \hline
     RetinexNet  & 38.56 & 23.90 & 35.47  &29.50 & 31.86 \\
     RetinexNet* & \underline{58.27} & 45.09 & \underline{57.84} & \underline{50.23} & \underline{52.86} \\ \hdashline
     & & & & & \\[-0.8em]
     URetinexNet  & 34.85 & 19.40 & 31.47 & 24.76 & 27.62 \\
     URetinexNet* & 56.75 & \underline{45.35} & 56.35 & 47.13 & 51.40 \\ \hdashline
     & & & & & \\[-0.8em]
     Ours & \textbf{69.45}  &\textbf{54.77} & \textbf{68.34} & \textbf{59.05} & \textbf{62.91}\\
    \hline
    \end{tabular}%
    }
  \label{ab:decomstrategy}%
\end{table}%

To further validate that the performance gains primarily originate from the latent decomposition strategy with a log transformation, we deploy it across various network architectures including RetinexNet~\cite{RetinexNet} and URetinexNet~\cite{uretinexnet}.
Concretely, for RetinexNet and URetinexNet, we retain original network architectures while only replacing the default pixel space image decoupling process with our proposed strategy, followed by retraining the image decomposition module.
As shown in Table~\ref{ab:decomstrategy}, after integrating our strategy, RetinexNet* and URetinexNet* achieve average performance gains of 21.00dB and 23.78dB, respectively, compared to their original counterparts.
Additionally, our full strategy attains superior performance owing to the transformer's stronger feature modeling capability.

\begin{figure}[!t]
    \centering
    \includegraphics[width=1.0\linewidth]{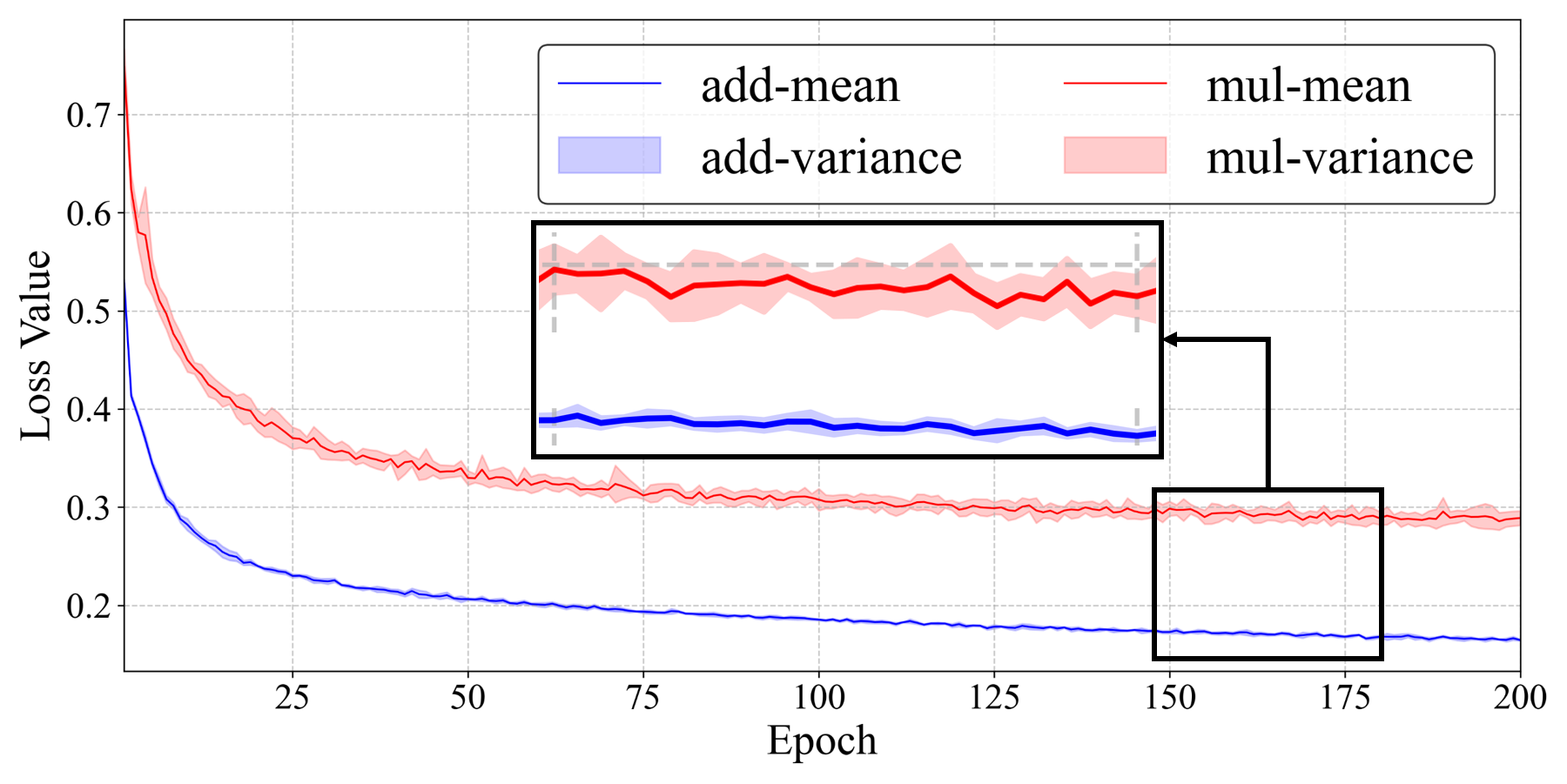}
    \caption{Comparison of training stability between additive and multiplicative decompositions. Each model is trained 10 times, and the figure shows the mean and variance of the loss at each epoch.}
    \label{fig:stable-10}
\end{figure}

In addition, to validate the improved training stability of the proposed decomposition strategy, both additive and multiplicative decomposition models are trained ten times. 
Statistical results of the enhancement PSNR show that the additive decomposition offers greater stability, with a mean PSNR of 23.84 and a variance of 0.067, whereas the multiplicative decomposition yields a mean PSNR of 22.16 and a variance of 0.157. 
The training loss curves in Fig.~\ref{fig:stable-10} further demonstrate that the additive decomposition attains a lower final convergence loss and converges faster. 
In addition, its loss curves display considerably smaller fluctuations across runs. These results indicate that under multiplicative decomposition, the values of $L$ and $R$ directly affect gradient propagation, introducing greater training uncertainty and inferior convergence.
In contrast, the proposed additive decomposition avoids this problem, leading to more stable training and more consistent enhancement performance.

\begin{figure}[!t]
    \centering
    \includegraphics[width=1.0\linewidth]{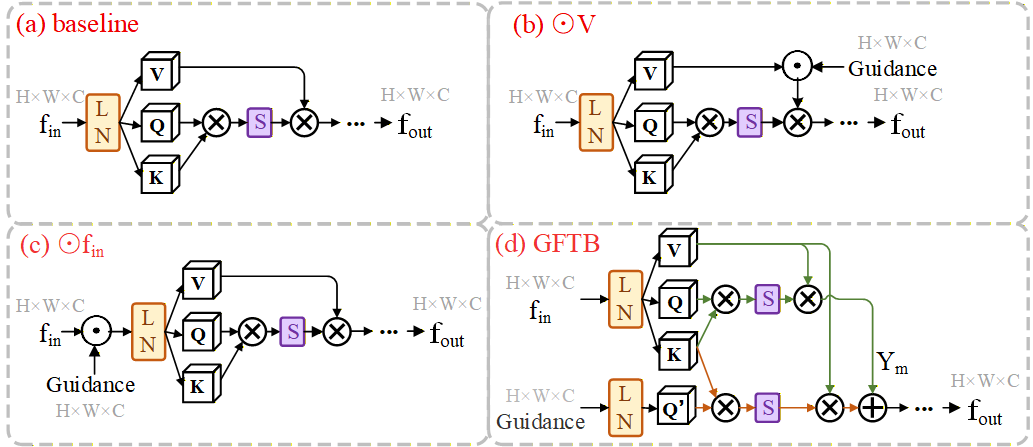}
    \caption{The overview of three GFTB variants and complete GFTB.}
    \label{fig:variants GFTB}
\end{figure}

\begin{table}[!t]
\renewcommand{\arraystretch}{1.20}
\centering
\caption{Ablation study on different GFTB structures.}
\label{ab-guidance}
\resizebox{1.0\linewidth}{!}{
\begin{tabular}{cccccccc|cc}
\hline
\multicolumn{8}{c|}{Different GFTB structures}    & \multicolumn{2}{c}{\multirow{2}{*}{Metrics}} \\ \cline{1-8}
\multicolumn{4}{c|}{Variants}    & \multicolumn{2}{c|}{R}            & \multicolumn{2}{c|}{L} & \multicolumn{2}{c}{}                         \\ \hline
\begin{tabular}[c]{@{}c@{}}(a)\\ SA\end{tabular} & \begin{tabular}[c]{@{}c@{}}(b)\\ $\odot V$\end{tabular} & \begin{tabular}[c]{@{}c@{}}(c)\\ $\odot f_{in}$\end{tabular} & \multicolumn{1}{c|}{\begin{tabular}[c]{@{}c@{}}(d)\\ CA\end{tabular}} & $S_{mean}$ & \multicolumn{1}{c|}{$S_{max}$} & $S_{mean}$      & $S_{max}$      & PSNR                  & SSIM                 \\ \hline
\checkmark  &   &     & \multicolumn{1}{c|}{}   &       & \multicolumn{1}{c|}{}     &            &           & 22.99                 & 0.837                \\\hdashline
& \checkmark &     & \multicolumn{1}{c|}{}   & \checkmark     & \multicolumn{1}{c|}{}     &            & \checkmark         & 23.03                 & 0.792                \\
&   & \checkmark   & \multicolumn{1}{c|}{}   & \checkmark     & \multicolumn{1}{l|}{}     &            & \checkmark         & 21.85                 & 0.815                \\
&   &     & \multicolumn{1}{l|}{\checkmark}  & \checkmark     & \multicolumn{1}{l|}{}     & \checkmark          &           & 23.53                 & 0.855                \\
&   &     & \multicolumn{1}{l|}{\checkmark}  &       & \multicolumn{1}{l|}{\checkmark}    &            & \checkmark         & 22.98                 & 0.842                \\
&   &     & \multicolumn{1}{l|}{\checkmark}  &       & \multicolumn{1}{l|}{\checkmark}    & \checkmark          &           & 23.36                 & 0.840                \\ \hdashline
&   &     & \multicolumn{1}{l|}{\checkmark}  & \checkmark     & \multicolumn{1}{l|}{}     &            & \checkmark         & \textbf{23.80}                 & \textbf{0.896}                \\ 
\hline
\end{tabular}
}
\end{table}

\textbf{Effectiveness of the guidance signals.}
To evaluate the contribution of the introduced guidance signals, we developed three GFTB variants, each incorporating a distinct guidance signal integration approach. As shown in Fig.~\ref{fig:variants GFTB}, (a) represents the GFTB without guidance signals. (b) and (c) correspond to GFTB configurations with two different feature fusion strategies. (d) denotes our proposed GFTB structure while allowing replacement of the guidance signal.

As indicated in Table~\ref{ab-guidance}, complete removal of guidance signals results in a 0.81 dB performance degradation of the model. 
Compared with the GFTB variants (\textit{i.e.}, $\odot V$ and $\odot f_{in}$), our proposed method achieves performance improvements of 0.78 dB and 1.95 dB over these two alternatives respectively. 
This demonstrates that the cross-attention based guidance signal integration approach is more effective. Additionally, compared with the full model, using only $S_{mean}$ or $S_{max}$ as the guidance signal leads to performance drops of 0.27 dB and 1.49 dB respectively. 
These results confirm that appropriate guidance signals are essential for achieving optimal enhancement performance.

\section{Conclusion}\label{Conclusions}
In this paper, we propose a Retinex-based low-light enhancement method, named Retinex-Guided Transformer~(RGT), featuring a two-stage architecture with decomposition and enhancement. 
Rooted in Retinex theory, we propose a latent-space decomposition strategy incorporating a log transformation and 1-pixel offset, achieving stable and accurate components of reflectance and illumination.
We propose a component refiner to enhance low-light images, where the designed guided feature transformer block preserves details of both reflectance and illumination components.
Comprehensive quantitative and qualitative experiments on four public datasets and real-world data demonstrate that RGT achieves superior performance. 
Ablation experiments validate the effectiveness of the proposed latent-space decomposition strategy.


 

%

\bibliographystyle{IEEEtran}
\bibliography{ref.bib}

\vfill

\end{document}